\documentclass[letterpaper]{article} 
\usepackage{aaai2026}  
\usepackage{times}  
\usepackage{helvet}  
\usepackage{courier}  
\usepackage[hyphens]{url}  
\usepackage{graphicx} 
\usepackage{natbib}  
\usepackage{caption} 
\usepackage{algorithm}
\usepackage{algorithmic}

\usepackage{newfloat}
\usepackage{listings}
\DeclareCaptionStyle{ruled}{labelfont=normalfont,labelsep=colon,strut=off} 
\floatstyle{ruled}
\newfloat{listing}{tb}{lst}{}
\floatname{listing}{Listing}
\usepackage{booktabs}
\usepackage{amsmath}
\usepackage{booktabs}  
\usepackage{multirow}
\usepackage{array}
\usepackage{tabularx}
\usepackage{amsthm}

\usepackage{pifont}
\newcommand{\cmark}{\ding{51}}  
\newcommand{\xmark}{\ding{55}}  

\title{From Hypothesis to Premises: LLM-based Backward Logical Reasoning with Selective Symbolic Translation}
\author{
    Qingchuan Li\textsuperscript{\rm 1}, Mingyue Cheng\textsuperscript{\rm 1}\thanks{Corresponding Author.}, Zirui Liu\textsuperscript{\rm 1}, Daoyu Wang\textsuperscript{\rm 1}, Yuting Zeng\textsuperscript{\rm 1}, Tongxuan Liu\textsuperscript{\rm 1, 2}
}
\affiliations{
    \textsuperscript{\rm 1}University of Science and Technology of China, 
    \textsuperscript{\rm 2}JD.com\\
    \{chouli, liuzirui, wdy030428, yuting\_zeng, tongxuan.ltx\}@mail.ustc.edu.cn, mycheng@ustc.edu.cn

}

\usepackage{bibentry}

\begin{document}

\maketitle

\begin{abstract}
Logical reasoning is a core challenge in natural language understanding and a fundamental capability of artificial intelligence, underpinning scientific discovery, mathematical theorem proving, and complex decision-making. Despite the remarkable progress of large language models (LLMs), most current approaches still rely on forward reasoning paradigms, generating step-by-step rationales from premises to conclusions. However, such methods often suffer from redundant inference paths, hallucinated steps, and semantic drift, resulting in inefficient and unreliable reasoning. In this paper, we propose a novel framework, Hypothesis-driven Backward Logical Reasoning (HBLR). The core idea is to integrate confidence-aware symbolic translation with hypothesis-driven backward reasoning. In the translation phase, only high-confidence spans are converted into logical form, such as First-Order Logic (FOL), while uncertain content remains in natural language. A translation reflection module further ensures semantic fidelity by evaluating symbolic outputs and reverting lossy ones back to text when necessary. In the reasoning phase, HBLR simulates human deductive thinking by assuming the conclusion is true and recursively verifying its premises. A reasoning reflection module further identifies and corrects flawed inference steps, enhancing logical coherence. Extensive experiments on five reasoning benchmarks demonstrate that HBLR consistently outperforms strong baselines in both accuracy and efficiency.
\end{abstract}

\begin{links}
    \link{Code}{https://github.com/wufeiwuwoshihua/HBLR}
\end{links}

\section{Introduction}

Logical reasoning lies at the heart of artificial intelligence (AI), playing a central role in scientific discovery, mathematical theorem proving, and complex decision-making~\cite{bronkhorst2020logical}. In natural language understanding, logical reasoning refers to the process of drawing valid conclusions from a set of textual premises, often requiring models to perform multi-step, structured inference~\cite{nunes2012logical, bronkhorst2020logical}. Despite its importance, reasoning over natural language remains a formidable challenge due to linguistic ambiguity, implicit knowledge, and the need for robust and reliable compositional generalization~\cite{ye2023satlm, pan2023logic, li2025llms, luo2025time}. Effective logical reasoning requires not only the understanding of individual statements, but also the ability to model their interrelations through valid and principled logical transformations~\cite{smith2003introduction}.


\begin{figure}[t]  
    \centering
    \includegraphics[width=0.95\linewidth]{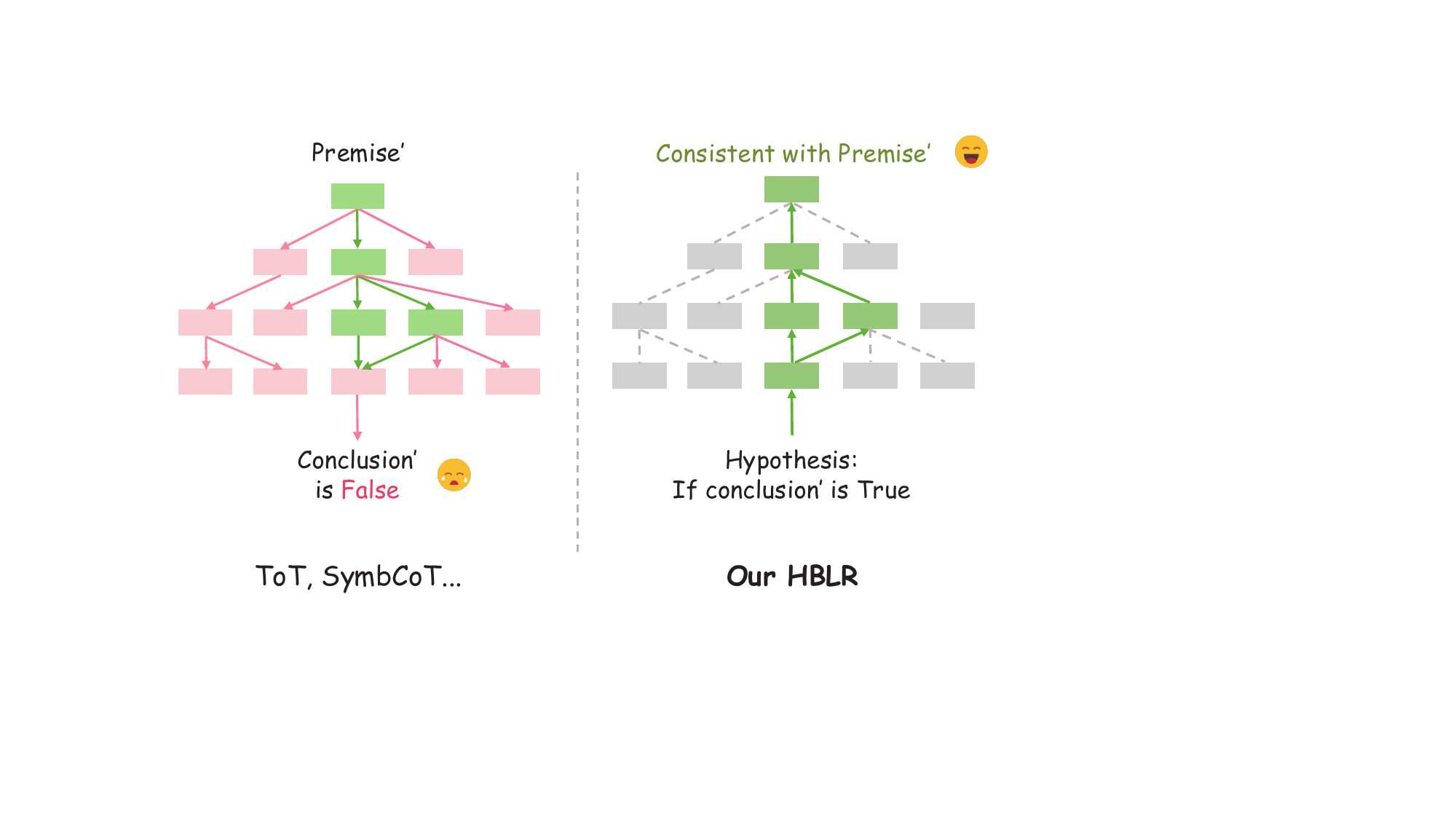 } 
\caption{
Conceptual comparison of reasoning paradigms. HBLR adopts a hypothesis-driven backward reasoning strategy with selective symbolic translation, enhancing precision and effectiveness.
}
\label{fig:intro}
\end{figure}

Over the past decades, a wide range of approaches have been proposed to tackle this challenge. Early methods relied extensively on symbolic representations and rule-based inference engines, such as first-order logic (FOL) systems and logic programming frameworks. Subsequently, constraint solvers (e.g., GeCode~\cite{schulte2006gecode}, PyKE~\cite{pyke}) and automated theorem provers (e.g., Prover9~\cite{prover9}, Z3~\cite{z3}) were introduced to strengthen formal inference capabilities. More recently, the emergence of neural-symbolic reasoning has enabled the integration of statistical learning with structured logical operations, bridging the gap between expressivity and scalability.

In particular, the emergence of large language models (LLMs)~\cite{patel2022bidirectional, hahn2022formal} has greatly propelled the development of natural language reasoning. A series of studies, such as CoT (Chain-of-Thought) prompting~\cite{wei2022cot}, Tree of Thoughts~\cite{yao2023tot}, LINC~\cite{olausson2023linc}, and Logic-LM~\cite{pan2023logic}, have demonstrated the potential of LLMs in performing complex deductive reasoning tasks. These methods either directly leverage LLMs to generate reasoning chains in natural language, or translate natural language into symbolic forms and invoke external solvers. Some recent works, such as SymbCoT~\cite{xu2024symbcot}, further integrate symbolic representations with LLMs to enhance reasoning faithfulness and interpretability.

Despite these promising advances, current methods still face several key limitations. First, most approaches adopt a forward reasoning paradigm, generating reasoning steps from premises toward conclusions. However, this paradigm often suffers from redundant reasoning paths, hallucinated intermediate steps, and goal deviation, resulting in unreliable or inefficient inference~\cite{kazemi_lambada_2023}. Second, symbolic translation is often applied globally to the entire input, despite the fact that large language models may only be confident about specific parts of the text. Over-reliance on imperfect translation may introduce logical errors or semantic loss~\cite{li2025llms}. Finally, reasoning steps are seldom verified or revised once generated, making the process prone to compounding mistakes.

To address these challenges, we propose a novel framework, Hypothesis-driven Backward Logical Reasoning (HBLR), shown in Figure \ref{fig:intro}. The core idea is to reframe logical reasoning as a hybrid process that (1) selectively translates high-confidence spans of natural language into formal logic and (2) simulates human-like deductive reasoning via backward chaining. Specifically, HBLR first performs confidence-aware symbolic translation into FOL while retaining uncertain parts in natural language. A translation reflection module further ensures semantic fidelity by reverting lossy translations. Then, HBLR assumes the conclusion to be true and recursively verifies its logical support in a hypothesis-driven, backward fashion. A reasoning reflection module is also introduced to detect and correct flawed inference steps, ensuring logical coherence and robustness.

\begin{itemize}
	\item We propose HBLR, a novel logical reasoning framework based on hypothesis-driven backward chaining, which simulates human deductive thinking by assuming the conclusion and verifying its premises in reverse, addressing the inefficiencies of forward reasoning paradigms.
	\item  We introduce a selective symbolic translation strategy that converts high-confidence spans into logic form while retaining uncertain parts in natural language, thereby balancing formal precision and semantic flexibility.
	\item 	We conduct comprehensive experiments on five benchmark datasets and show that HBLR achieves superior performance in reasoning accuracy and efficiency.

\end{itemize}

\section{Related Work}

\paragraph{Prompt-based LLM Reasoning.} Logical reasoning—deriving correct conclusions from given premises—is a core ability for large language models (LLMs)~\citep{mondorf_survey_2024,sun_survey_2024,qiao_reasoning_2023}. Prompting provides a direct way to activate this ability. For complex tasks, step-by-step prompting has become widely used~\citep{besta_topologies_2024,wang_self-consistency_2023}. Chain-of-Thought (CoT) prompting~\citep{wei2022cot} guides LLMs to generate intermediate steps, while Tree-of-Thought (ToT)~\citep{yao2023tot} and Graph-of-Thought (GoT)~\citep{besta_graph_2024} enable exploration of multiple reasoning paths. However, these methods still struggle in non-mathematical tasks and when exemplar and target question complexity differ~\citep{sprague2024cotcotchainofthoughthelps}.

Another major direction is question decomposition~\citep{zhang_cumulative_2023,yao_react_2023,kazemi_lambada_2023}. Least-to-most prompting~\citep{zhou_least--most_2023, wang2025can, jiang2025tablemind} solves problems by first handling simpler sub-questions, and divide-and-conquer strategies~\citep{cui2023a,zhang2024exam,cheng2025survey} further improve consistency and logical accuracy. Backward chaining~\citep{kazemi_lambada_2023} reduces the search space by breaking tasks into solvable sub-modules. Despite these advances, prompt-based reasoning still faces high computational cost and unstable performance~\citep{yao2023tot}.

\paragraph{Symbolic-based LLM Reasoning.} Symbolic reasoning uses formal logic symbols and expressions to enable consistent and precise inference, helping reduce issues in prompt-based reasoning such as inconsistency and sensitivity to premise order~\citep{chen_premise_2024,bao_llms_2024}.

One main line of work improves LLM reasoning by introducing explicit logical representations~\citep{wang_logic-driven_2022,wan__2024}. \citet{wang_logic-driven_2022} mapped natural language to logical forms to produce more aligned and reliable answers, while \citet{bao-etal-2024-abstract} used structured semantic graphs for logic-driven data augmentation across diverse tasks. \citet{xu2024symbcot} proposed a two-stage method that first generates logical forms and then uses them to guide downstream reasoning and planning more effectively.
Another direction employs symbolic solvers to infer over LLM-generated logic~\citep{olausson2023linc,pan2023logic,ye2023satlm}. Solver choice, such as SAT~\citep{ye2023satlm} or first-order logic~\citep{pan2023logic}, directly affects accuracy and overall generalization quality. SymBa~\citep{lee2024symba} further integrates classical SLD resolution with LLMs, providing symbolically guided chain-of-thought reasoning that significantly improves both performance and interpretability.

Despite these strengths, symbolic methods remain limited by the translation step from natural language to logic: errors or missing information in this process can weaken downstream reasoning~\citep{pan2023logic}.

\section{Preliminaries}

\subsection{Problem Definition}

We study the task of natural language logical reasoning, where the input consists of a set of premises $\mathcal{P}$ and a target conclusion $\mathcal{C}$. The goal is to determine whether the conclusion can be logically inferred from the premises, denoted as $\mathcal{P} \models \mathcal{C}$, meaning that $\mathcal{C}$ is true in all possible interpretations where $\mathcal{P}$ holds. In our setting, both $\mathcal{P}$ and $\mathcal{C}$ may contain a mixture of formal logic expressions and natural language statements. After symbolic translation, the inputs are reformulated as a hybrid premise set $\mathcal{P'}$ and a hybrid conclusion $\mathcal{C'}$. The final goal is to determine whether $\mathcal{P'} \models \mathcal{C'}$.

\subsection{Empirical Exploration}

\begin{table}[t]
\centering
{
\fontsize{9}{10}\selectfont
\begin{tabular}{lccc}
\toprule
\textbf{Dataset} & \textbf{Solver} & \textbf{GPT-4} & \textbf{DeepSeek-V3} \\
\midrule
FOLIO & Prover9 & 0.7383 & 0.7245 \\
ProofWriter & PyKE & 0.8338 & 0.8211 \\
ProntoQA & PyKE & 0.9050 & 0.8783 \\
Deduction & constraint & 0.9599 & 0.8963 \\
AR-LSAT & Z3 & 0.4304 & 0.4041 \\
\bottomrule
\end{tabular}
}
\caption{ 
Datasets, their associated symbolic solvers and the translation accuracy of GPT-4 and DeepSeek-V3.  Note: ``Deduction'' denotes the LogicalDeduction dataset, and ``constraint'' refers to the python-constraint solver.
}
\label{tab:solver_accuracy_combined}
\end{table}

\paragraph{Translation Module.}

We first evaluate the SymbCoT~\cite{xu2024symbcot} translation module to measure how well LLMs convert natural language into formal logic. We keep the original SymbCoT translator but replace its LLM-based reasoning with symbolic solvers, using solver outputs as proxies for translation accuracy within well-defined logical domains. Experiments are run on five datasets~\cite{saparov2022prontoqa, tafjord2020proofwriter, ghazal2013deduction, han2022folio, zhong2021ar} and four symbolic solvers, with GPT-4 and DeepSeek-V3 as base models, following the setup of Logic-LM~\cite{pan2023logic}. As shown in Table~\ref{tab:solver_accuracy_combined}, LLMs achieve high translation accuracy on synthetic datasets like ProntoQA and ProofWriter, which use clear logical templates. However, performance drops sharply on manually curated datasets such as FOLIO and AR-LSAT, where natural language is more varied and logical structure is often implicit. These results show that LLMs handle explicit logical inputs well but struggle when logic is expressed indirectly. This motivates a selective translation strategy that converts only high-confidence spans to logic and keeps ambiguous parts in natural language to reduce semantic drift.

\begin{figure}[t]  
    \centering
    \includegraphics[width=1\linewidth]{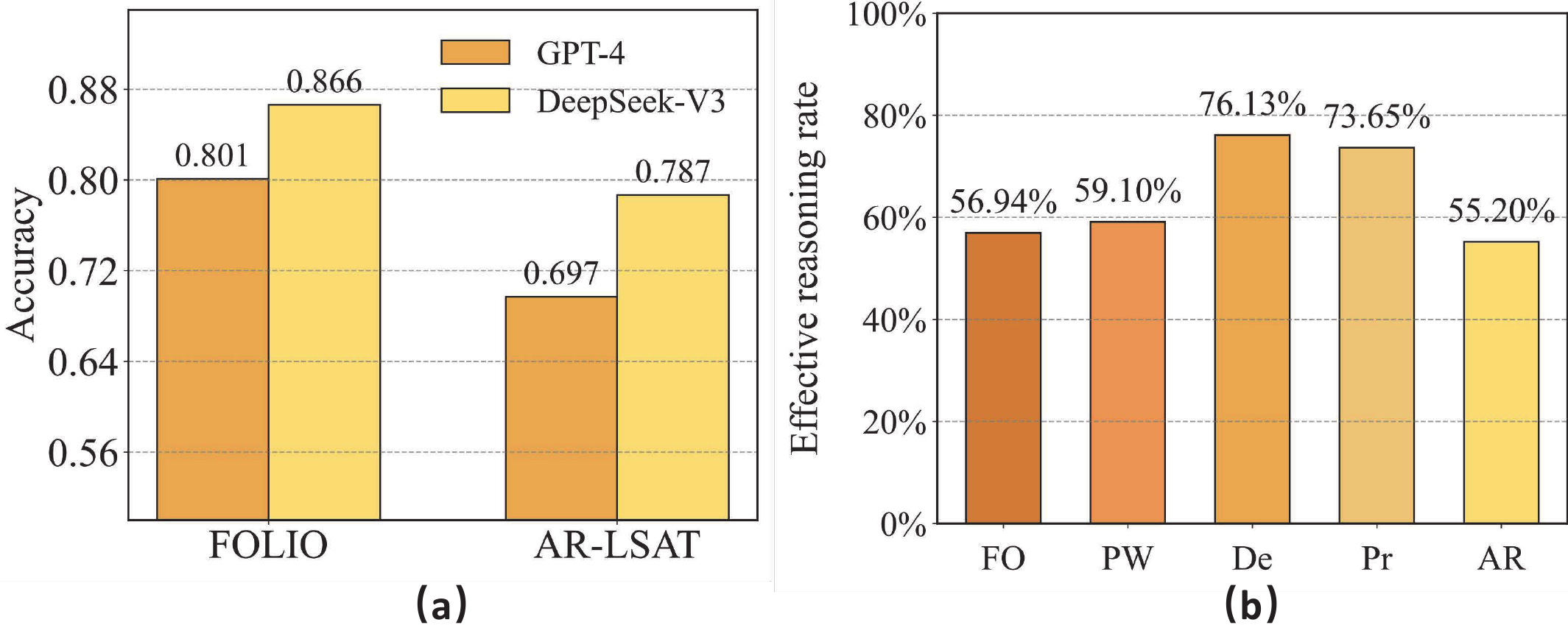 }
\caption{
(a) Translation Error Rate Among SymbCoT Failure Cases in FOLIO and AR-LSAT 
(b) Ratio of essential tokens retained after pruning redundant steps from GPT-4-generated reasoning plans. 
}

    \label{fig:token_ratio}
\end{figure}

To evaluate how translation errors affect SymbCoT’s LLM-based reasoning module, we examine two datasets with low translation accuracy, FOLIO and AR-LSAT, and measure the share of reasoning failures linked to translation errors. As shown in Figure~\ref{fig:token_ratio}a, these errors are a major source of SymbCoT’s reasoning failures on both datasets.

\paragraph{Reasoning Module.}

We further analyze the efficiency of SymbCoT’s reasoning strategy. Although SymbCoT employs a plan-and-solve approach that generates reasoning chains from premises to conclusions, we observe that its LLM-based reasoning often lacks goal-directedness—frequently invoking irrelevant premises and introducing redundant steps. To quantify this inefficiency, we extract essential reasoning paths from correct GPT-4 predictions by pruning extraneous steps while retaining all necessary inferences. We then compute the ratio of tokens in the essential paths relative to the original plans (Figure~\ref{fig:token_ratio}b). The results show that forward reasoning tends to overuse available premises, resulting in unnecessarily verbose inference trajectories. These findings highlight the need for a goal-driven alternative that begins with the hypothesis and selectively identifies minimal supporting premises, thereby improving reasoning efficiency and precision.

\begin{figure*}[t] 
    \centering
    \includegraphics[width=0.98\textwidth]{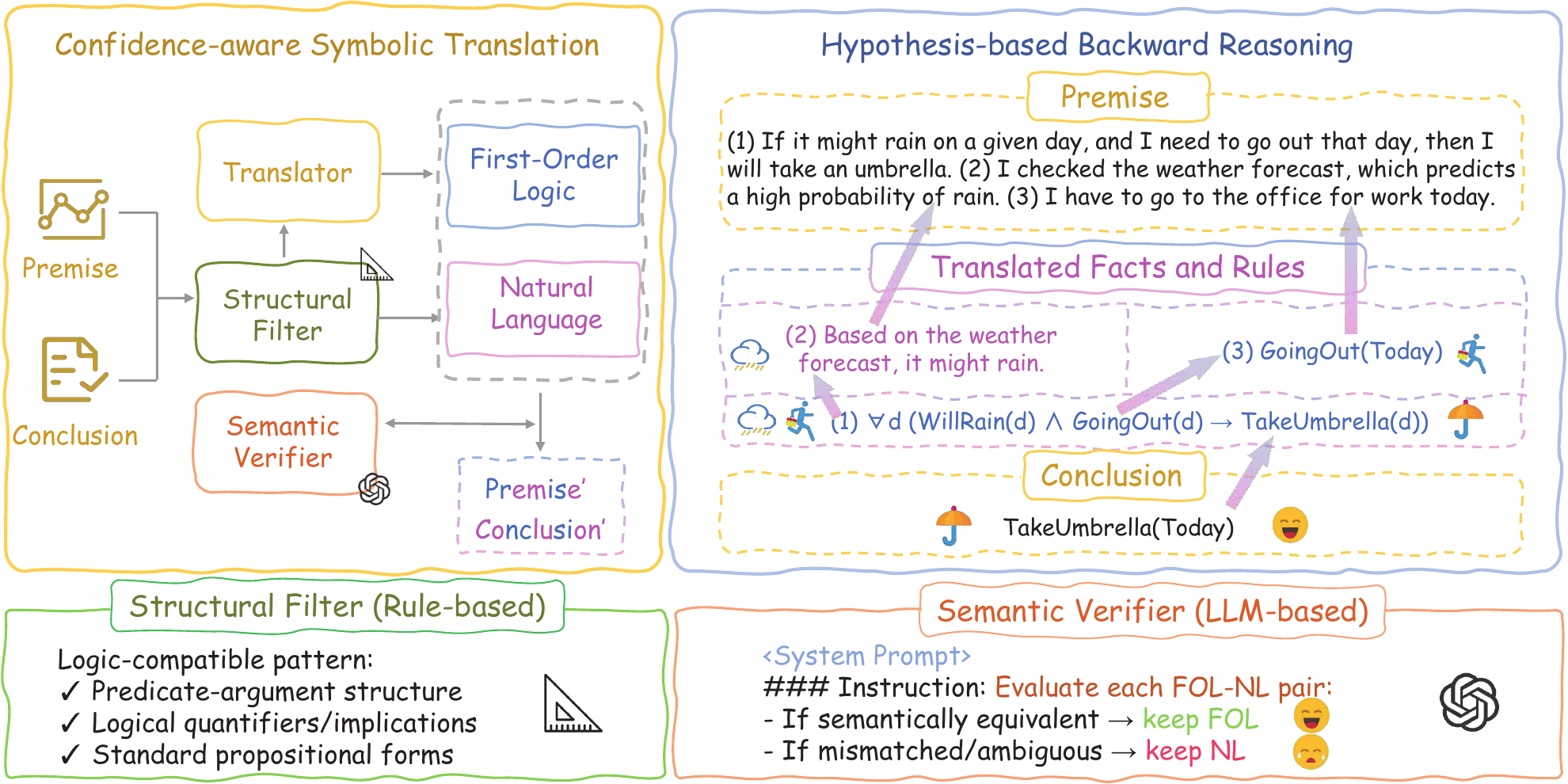} 
    \caption{Overview of the Hypothesis-driven Backward Logical Reasoning (HBLR) framework. Structural Filter and Semantic Verifier below illustrate the internal mechanisms of the Confidence-aware Symbolic Translation module.}
    \label{fig:method}
\end{figure*}
\section{Methodology}

\subsection{Overview of the HBLR Framework}

To overcome the limitations of existing methods in symbolic translation and reasoning, we propose Hypothesis-driven Backward Logical Reasoning (HBLR) (Figure~\ref{fig:method}). HBLR improves translation reliability and strengthens goal-directed reasoning through two components: the Confidence-aware Symbolic Translation Module (CSTM) and the Hypothesis-based Backward Reasoning Module (HBRM). CSTM selectively converts natural language into formal logic based on structural and semantic confidence, while HBRM mirrors human deductive reasoning by starting from the conclusion and working backward to find the minimal supporting premises.

\subsection{Confidence-aware Symbolic Translation}

To reduce errors from full symbolic translation, we introduce the Confidence-aware Symbolic Translation Module (CSTM), which selectively converts structurally sound and semantically clear statements into logical forms while keeping the rest in natural language. This hybrid strategy balances the rigor of symbolic logic with the expressiveness and robustness of natural language.
CSTM uses two mechanisms—a structural rule-based pre-checker and a semantic consistency verifier—to ensure both syntactic and semantic requirements are met before translation.

\paragraph{Structural Filter.}

To determine whether a sentence $s$ is suitable for symbolic translation, we apply a structural filter that detects logic-compatible patterns in natural language. Sentences that pass this filter are translated into a logical form $\phi$, while those that do not are retained in natural language to preserve semantic fidelity.

We define logic-compatible patterns as a specific class of linguistic structures that are suitable for formal representation. A sentence is considered logic-compatible if it conforms to these predefined patterns, which include explicit predicate-argument constructions, the presence of logical connectives or quantifiers such as “if,” “then,” “and,” or “or,” and canonical formulations commonly found in propositional or first-order logic. These structural characteristics ensure that the sentence can be reliably and meaningfully translated into symbolic form.

\paragraph{Semantic Verifier.}

To ensure that symbolic translations faithfully preserve the meaning of the original sentence, we introduce a semantic verifier. For each candidate logical form $\phi$ generated from a natural language sentence $s$, the verifier assesses whether $\phi$ semantically aligns with the intent of $s$. This check is performed using LLM-based prompting with carefully selected few-shot exemplars.

The verifier is designed as a conservative safeguard rather than a perfect semantic judge. Because an incorrect logical form is more harmful than retaining the original natural language, the system adopts a strict acceptance rule: $\phi$ is accepted only when it can be confidently verified as fully consistent with $s$. Any logical form with uncertainty or insufficient semantic evidence is rejected, and the corresponding $s$ is preserved in natural language. This conservative strategy reduces semantic drift and ensures that symbolic representations remain faithful and reliable, avoiding the introduction of incorrect logic when translation is uncertain.

\paragraph{Hybrid Representation.}
The final output of CSTM consists of two complementary components: a set of high-confidence logical expressions $\{\phi_i\}_{i=1}^m$ and a set of natural language statements $\{s_j\}_{j=1}^n$, which are retained due to insufficient structural or semantic confidence for symbolic translation. These elements are integrated into a unified hybrid context that serves as the input for downstream reasoning. Formally, we define the hybrid context as:
\begin{equation}
\text{HybridContext} = \mathcal{P'} \cup {\mathcal{C'}}
\end{equation}
where $\mathcal{P'} = \{ \phi_i \} \cup \{ s_j \}$ denotes the premise set, comprising both logical expressions and natural language statements, and $\mathcal{C'} = \{\phi_m\} \cup \{s_n\}$ represents the conclusion, expressed in logical or textual form depending on translation confidence. This hybrid representation enables flexible yet faithful reasoning by combining the precision of formal logic with the semantic richness of natural language.

\subsection{Hypothesis-based Backward Reasoning}

To address the inefficiency and lack of clear goal orientation inherent in conventional forward reasoning approaches, we introduce the Hypothesis-based Backward Reasoning Module (HBRM). Inspired by the classical hypothetical-deductive paradigm, HBRM initiates reasoning by assuming the conclusion to be true and recursively verifying the premises that would logically support it.

\paragraph{Backward Reasoning with Hypothesis.}

The input to HBRM is defined as a tuple $(\mathcal{P'}, \mathcal{H})$, where $\mathcal{P'}$ denotes the hybrid premise set, containing both logical expressions and natural language statements, and $\mathcal{H}$ is the hypothesis explicitly asserting that the target conclusion $\mathcal{C}$ holds true. Reasoning unfolds by constructing a sequence of intermediate hypotheses $\{H_t\}$, each of which is iteratively evaluated through deductive inference based on the information encoded in $P$. This backward chaining process is formally summarized in Algorithm~\ref{deduc}.

\begin{algorithm}
\caption{Backward Reasoning with Hypothesis}
\label{deduc}
\begin{algorithmic}[1]
\REQUIRE Hypothesis $\mathcal{H}$, Hybrid Premises $\mathcal{P'}$
\ENSURE Validity status of $\mathcal{H}$

\WHILE{steps $\leq$ $k$}
    \STATE $Z \gets$ \textsc{Reasoning}($\mathcal{P'}$, $\mathcal{H}$)
    \IF{$Z$ contradicts $\mathcal{P'}$ or $\mathcal{H}$}
        \STATE \textbf{return} False
    \ELSIF{$Z$ supports $\mathcal{P'}$ or $\mathcal{H}$}
        \STATE \textbf{return} True
    \ELSE
        \STATE $\mathcal{H} \gets Z$
    \ENDIF
\ENDWHILE
\STATE \textbf{return} Unknown
\end{algorithmic}
\end{algorithm}

\begin{table}[t]
\centering
\setlength{\tabcolsep}{1mm}
{%
\fontsize{9}{10}\selectfont
\begin{tabular}{lcccc}
\toprule
\textbf{Dimension} & \textbf{CoT} & \textbf{Logic-LM} & \textbf{SymbCoT} & \textbf{HBLR} \\
\midrule
Use of Solver & \xmark & \cmark & \xmark & \xmark \\
Translation & None & Full & Full & Selective \\
Strategy & Forward & Solver & Forward & Backward \\
Redundancy & High & -- & High & Low \\
Interpretability & High & Low & High & High \\
Verification & No & No & Yes & Yes \\
\bottomrule
\end{tabular}
}
\caption{Comparison of HBLR and baseline methods. HBLR does not rely on external solvers and achieves low reasoning redundancy and high interpretability through backward reasoning with confidence-aware symbolic translation.}
\label{tab:method_comparison}
\end{table}

\paragraph{Stopping Criteria.}

The backward reasoning process terminates when any of the following conditions is met: (i) the current hypothesis $H_t$ is directly entailed by a premise in $P$, indicating a successful and complete proof; (ii) $H_t$ contradicts an existing premise in $P$, resulting in a logical refutation; or (iii) the maximum number of reasoning steps is reached without sufficient supporting evidence, in which case the outcome is deemed logically inconclusive.

\paragraph{Verification Mechanism.}
To ensure logical soundness and semantic fidelity, HBRM incorporates a verification mechanism that evaluates the validity of each reasoning step. If a step contains logical errors, semantic inconsistencies, or unsupported inferences, the system reconstructs a revised reasoning path; otherwise, the original path is preserved.
By guiding the inference process in a backward manner, from the conclusion toward its underlying premises, HBRM enhances goal alignment, reduces redundancy, and produces more efficient and interpretable reasoning trajectories.

\subsection{Discussions}

As shown in Table~\ref{tab:method_comparison}, HBLR addresses several limitations of existing reasoning frameworks. Unlike Logic-LM and SymbCoT, which fully translate all inputs into symbolic logic, HBLR performs selective symbolic translation, converting only high-confidence spans while retaining natural language when appropriate. This reduces translation errors and better preserves contextual semantics.
In contrast to CoT and SymbCoT, which rely on forward reasoning, HBLR employs a hypothesis-driven backward reasoning mechanism that initiates inference from the conclusion and recursively identifies minimal supporting premises. This goal-oriented approach significantly reduces redundancy in reasoning paths.
HBLR also incorporates a step-wise verification mechanism that evaluates each inference step, improving both reliability and interpretability. Overall, HBLR combines the precision of symbolic methods with the flexibility of neural reasoning, while mitigating the weaknesses of both.

\section{Experiments}

\subsection{Experimental Settings}

\paragraph{Evaluation Models.}
Experiments are conducted using four representative LLMs: the relatively less capable \textbf{GPT-3.5-Turbo}  \cite{openai_gpt35}; the more advanced \textbf{GPT-4}  \cite{achiam2023gpt4} and \textbf{DeepSeek-V3}  \cite{liu2024deepseekv3}; and \textbf{DeepSeek-R1}  \cite{guo2025deepseekr1}, currently one of the most reasoning-capable models available.

\paragraph{Evaluation Datasets.}
Our evaluation spans five widely-used logical reasoning benchmarks: \textbf{ProntoQA} \cite{saparov2022prontoqa}, \textbf{ProofWriter} \cite{tafjord2020proofwriter}, \textbf{FOLIO} \cite{han2022folio}, \textbf{LogicalDeduction} \cite{ghazal2013deduction}, and \textbf{AR-LSAT}~ \cite{zhong2021ar}. These datasets vary in symbolic formalisms and present diverse challenges across deductive reasoning scenarios. We adopt \textbf{accuracy} as the primary evaluation metric, measuring the correctness of multiple-choice answers.

\paragraph{Symbolic Formalisms.}
For ProntoQA, ProofWriter, and FOLIO, we use first-order logic (FOL) as the primary underlying symbolic structure. To assess the broader generalizability of our approach, we additionally evaluate on constraint optimization (CO) symbolic representations in LogicalDeduction and AR-LSAT.

\paragraph{Baselines.}
We compare HBLR against several strong baselines employing distinct strategies:
(1) \textbf{Direct}, which uses LLMs to directly answer questions without intermediate reasoning;
(2) \textbf{CoT} \cite{wei2022cot}, which applies chain-of-thought prompting to elicit step-by-step reasoning;
(3) \textbf{Logic-LM} \cite{pan2023logic}, which translates problems into logic and invokes symbolic solvers; and
(4) \textbf{SymbCoT}~ \cite{xu2024symbcot}, which combines symbolic translation with LLM-based reasoning.

\begin{table*}[t]
\centering
{
\fontsize{9}{10}\selectfont
\begin{tabular}{lccccc|ccccc}
\toprule
\multirow{2}{*}{\textbf{Dataset}} 
& \multicolumn{5}{c|}{\textbf{GPT-4}} 
& \multicolumn{5}{c}{\textbf{GPT-3.5-Turbo}} \\
\cmidrule(lr){2-6} \cmidrule(lr){7-11}
& Direct & CoT & Logic & SymbCoT & HBLR 
& Direct & CoT & Logic & SymbCoT & HBLR \\
\midrule
\textbf{ProntoQA}    & 77.40 & \underline{94.79} & 90.50 & 97.16 & \textbf{99.36} & 46.04 & 67.80 & 67.21 & \underline{71.95} & \textbf{75.58} \\
\textbf{ProofWriter} & 52.67 & 68.11 & \underline{83.38} & 79.34 & \textbf{89.41} & 36.53 & 49.17 & 58.62 & \underline{59.03} & \textbf{63.24} \\
\textbf{FOLIO}       & 70.61 & 72.37 & 73.83 & \underline{78.19} & \textbf{84.22} & 45.09 & 57.35 & 48.83 & \underline{57.84} & \textbf{59.12} \\
\textbf{Deduction}   & 71.33 & 75.25 & \underline{95.99} & 93.00 & \textbf{97.83} & 39.15 & 43.67 & \textbf{69.06} & 45.85 & \underline{49.77} \\
\textbf{AR-LSAT}     & 34.43 & 35.06 & \underline{43.04} & 37.19 & \textbf{44.67} & 20.34 & 21.31 & \underline{25.15} & 21.59 & \textbf{27.22} \\
\midrule
\multicolumn{11}{c}{} \\[-1.2em]
\toprule
\multirow{2}{*}{\textbf{Dataset}} 
& \multicolumn{5}{c|}{\textbf{DeepSeek-V3}} 
& \multicolumn{5}{c}{\textbf{DeepSeek-R1}} \\
\cmidrule(lr){2-6} \cmidrule(lr){7-11}
& Direct & CoT & Logic & SymbCoT & HBLR 
& Direct & CoT & Logic & SymbCoT & HBLR \\
\midrule
\textbf{ProntoQA}    & 74.93 & 97.67 & 87.83 & \underline{98.43} & \textbf{99.55} & 97.28 & \underline{99.57} & 84.21 & 98.47 & \textbf{99.72} \\
\textbf{ProofWriter} & 54.46 & 56.84 & 82.11 & \underline{84.15} & \textbf{89.48} & 82.48 & 86.27 & 84.35 & \underline{88.34} & \textbf{92.32} \\
\textbf{FOLIO}       & 68.36 & 74.64 & 72.45 & \underline{77.78} & \textbf{85.22} & 88.13 & \underline{92.97} & 72.51 & 84.46 & \textbf{95.60} \\
\textbf{Deduction}   & 67.25 & 72.67 & \underline{90.63} & 89.91 & \textbf{92.63} & 76.17 & 86.45 & \underline{99.50} & 99.03 & \textbf{99.61} \\
\textbf{AR-LSAT}     & 36.21 & 42.50 & 40.41 & \underline{44.52} & \textbf{46.69} & 60.40 & \underline{76.91} & 62.10 & 70.87 & \textbf{86.47} \\
\bottomrule
\end{tabular}
}
\caption{Performance Comparison between HBLR and Baselines on Logical Reasoning Datasets. The second-best score is \underline{underlined} and \textbf{bold} one is the best.}
\label{table:main}
\end{table*}

\subsection{Overall Evaluation Results}
As shown in Table~\ref{table:main}, HBLR consistently outperforms all baselines across five benchmark datasets. On GPT-4, it achieves gains of up to 36.74\%, 22.58\%, 10.39\%, and 10.07\% over Direct, CoT, Logic-LM, and SymbCoT, respectively. Similar trends are observed on GPT-3.5-Turbo, DeepSeek-V3, and DeepSeek-R1, confirming the robustness of HBLR across models with varying reasoning capabilities.
Compared to Logic-LM and SymbCoT, the improvements highlight the effectiveness of HBLR’s partial symbolic translation and backward reasoning in balancing logical formality with LLM-native reasoning. Notably, HBLR outperforms Logic-LM by an average of 14.61\% on DeepSeek-R1.
An exception occurs on the LogicalDeduction task with GPT-3.5-Turbo, where Logic-LM slightly outperforms HBLR. This is due to GPT-3.5-Turbo’s limited reasoning ability—while it handles translation reasonably well, it struggles with complex inference. Logic-LM avoids this bottleneck by delegating reasoning to an external solver.

Another notable trend is that Logic-LM and SymbCoT often rank second-best on GPT-4, GPT-3.5-Turbo, and DeepSeek-V3, while CoT ranks second on DeepSeek-R1. This suggests that as model reasoning improves, fully symbolic approaches may introduce noise that offsets their benefits. HBLR remains robust by minimizing such noise while leveraging stronger reasoning capacity.

\subsection{Impact of Translation Module}
\begin{figure}
    \centering
    \includegraphics[width=0.95\linewidth]{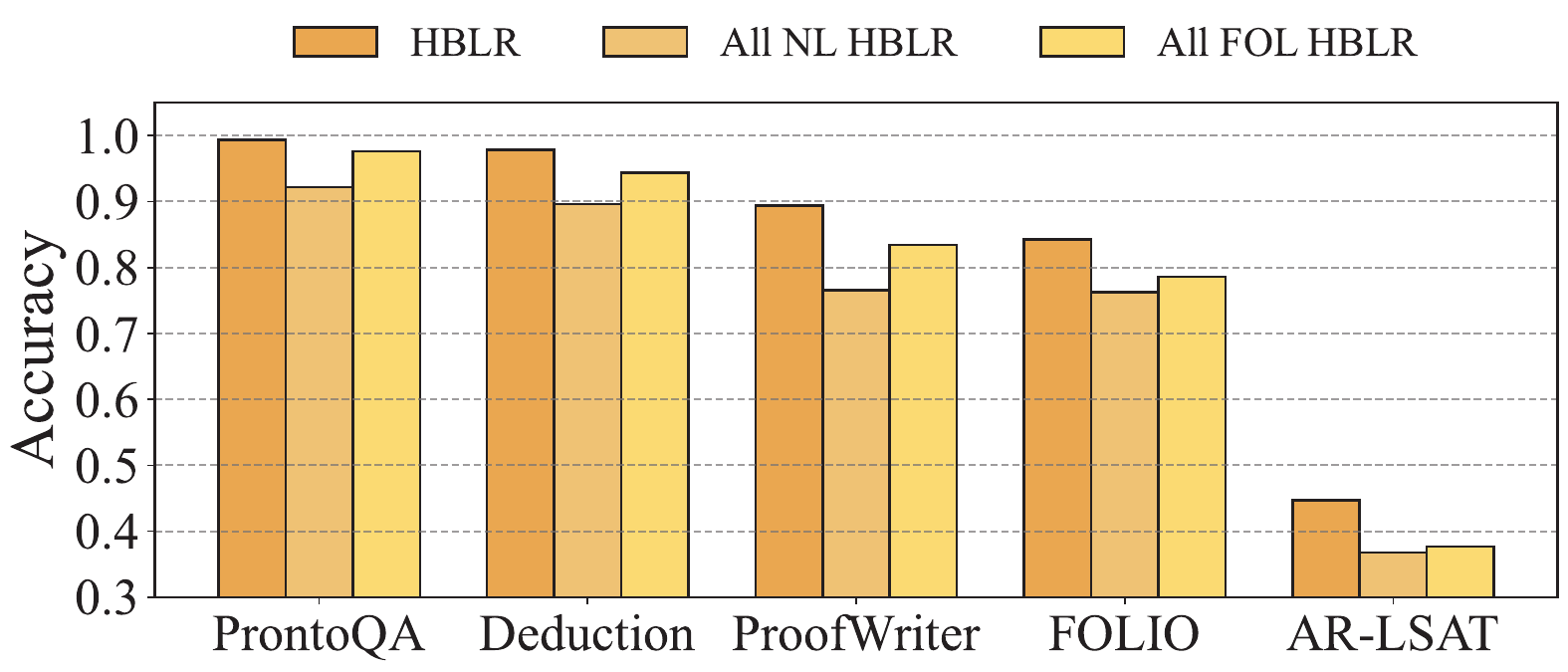}
    \caption{Comparison of selective translation (HBLR) with its two module variants: All-NL and All-FOL. HBLR consistently achieves higher accuracy across all datasets by balancing natural language and formal logic.}
    \label{fig:translation_ablation}
\end{figure}

To assess the effectiveness of our selective translation strategy, we compare it with two ablated variants of the HBLR translation module: \textbf{All-NL}, which performs no translation and retains all inputs in natural language; and \textbf{All-FOL}, which fully translates all inputs into formal logic. As shown in Figure~\ref{fig:translation_ablation}, our selective strategy consistently outperforms both variants across all five benchmarks.

On average, the selective strategy yields a 7.2\% improvement over All-NL, showing that natural language alone lacks sufficient structure for accurate reasoning. It also outperforms All-FOL by 4.7\%, indicating that full formalization can introduce translation errors or add unnecessary rigidity. The gains are especially large on \textbf{ProofWriter} (+12.83\%) and \textbf{LogicalDeduction} (+8.21\%), both requiring multi-step reasoning. In contrast, \textbf{ProntoQA} shows a small decline (-1.72\%) compared to All-FOL, likely because its highly structured format benefits from full logical conversion. Overall, these results show that the selective strategy effectively combines the semantic richness of natural language with the structural precision of formal logic, allowing LLMs to adapt to task demands and input characteristics.
\begin{figure}
    \centering
    \includegraphics[width=1\linewidth]{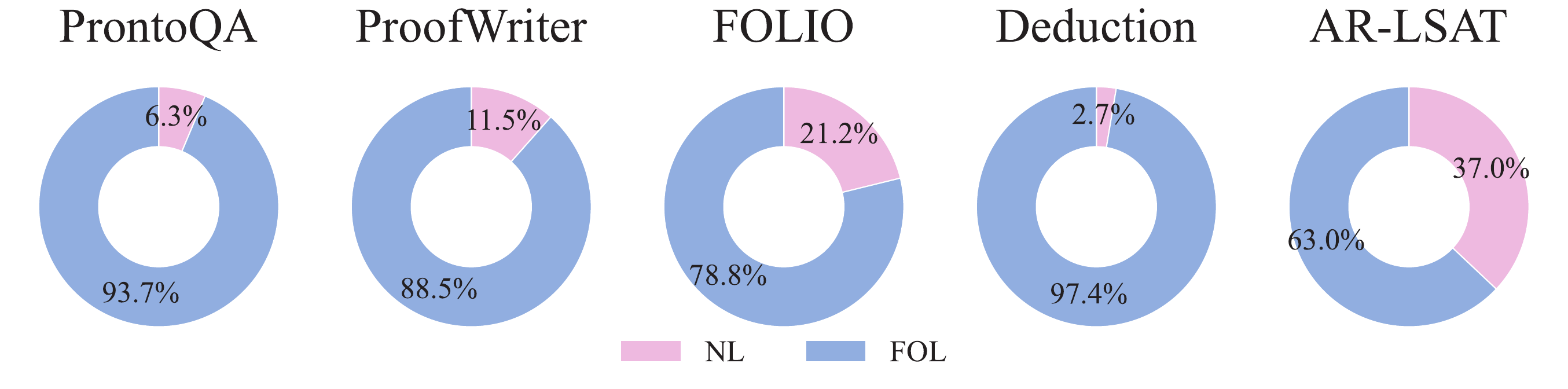}
    \caption{Proportion of natural language retained by the Translation module across datasets. Retention varies by task, reflecting differences in translation confidence.}
    \label{fig:nl_ratio}
\end{figure}

We analyze the proportion of natural language retained by HBLR’s translation module using GPT-4 (Figure~\ref{fig:nl_ratio}). On average, only 15.74\% of the input remains in natural language, with the lowest retention on \textbf{LogicalDeduction} (2.65\%), showing that most inputs are confidently translated into formal logic. This matches our confidence-aware design: natural language is kept only for uncertain segments, reducing ambiguity while limiting translation errors.
Higher retention on \textbf{AR-LSAT} (37.03\%) and \textbf{FOLIO} (21.17\%) aligns with Section~3.2, where we show these datasets have greater translation uncertainty. These results confirm that the selective strategy adjusts to dataset-specific reliability and preserves robustness when symbolic translation is less certain.

\subsection{Impact of Reasoning Module}

\begin{table}[t]
\centering
{
\fontsize{9}{10}\selectfont
\begin{tabular}{lccc}
\toprule
\textbf{Dataset} & \textbf{HBLR} (\%) & \textbf{Fwd Var.} (\%) & \textbf{$\Delta$} \\
\midrule
ProntoQA & 99.36 & 95.41 & +3.95 \\
ProofWriter & 89.41 & 81.24 & +8.17 \\
FOLIO & 84.22 & 77.58 & +6.64 \\
Deduction & 97.83 & 93.34 & +4.49 \\
AR-LSAT & 44.07 & 38.86 & +5.21 \\
\bottomrule
\end{tabular}
}
\caption{Comparison between HBLR and its forward reasoning variant (Fwd Var.) across five datasets. HBLR consistently achieves higher accuracy.}
\label{tab:reasoning_comparison}
\end{table}

To evaluate the contribution of our HBRM strategy, we compare it against a forward reasoning variant that adopts CoT prompting, while keeping the verification module unchanged. As shown in Table~\ref{tab:reasoning_comparison}, HBLR consistently outperforms CoT across all five datasets.

On average, backward reasoning yields a 5.69\% improvement. This gain can be explained as follows: forward reasoning tends to expand redundant branches during the search and accumulate translation errors, whereas HBLR starts from the hypothesis and traces back the key premises that support it, which greatly reduces the search space and limits error propagation. As a result, HBLR produces shorter and more accurate reasoning chains. This advantage is especially clear on ProofWriter (+8.17\%), where long reasoning chains make forward reasoning prone to early errors that amplify later; HBLR’s backward strategy focuses on essential premises more quickly and keeps the reasoning process stable. On AR-LSAT, which involves complex structures and diverse semantics, HBLR still achieves a +5.21\% gain, demonstrating stronger robustness and generalization. Overall, hypothesis-driven backward reasoning effectively controls the direction of inference, reduces noise and redundancy, and leads to more reliable reasoning across tasks.

\begin{table}[t]
\centering
{
\fontsize{9}{10}\selectfont
\begin{tabular}{lccc}
\toprule
\textbf{Dataset} & \textbf{SymbCoT} (\%) & \textbf{HBLR} (\%) & \textbf{$\Delta$}  \\
\midrule
ProntoQA & 73.65 & 95.58 & 21.93\\
ProofWriter & 59.10 & 81.60 & 22.50 \\
FOLIO & 56.94 & 73.29 & 16.35  \\
Deduction & 76.13 & 87.46 & 11.33  \\
AR-LSAT & 55.20 & 66.08 & 10.88  \\
\bottomrule
\end{tabular}
}
\caption{Effective reasoning rates of SymbCoT and HBLR. HBLR shows consistent improvements across datasets.}
\end{table}

Following the methodology in Section~3.2, we compute the proportion of effective reasoning steps produced by HBLR using GPT-4 and compare it with the previously reported SymbCoT results. HBLR substantially increases the share of effective reasoning, with improvements up to 22.50\%. It reaches over 70\% effectiveness on all datasets except \textbf{AR-LSAT}, which contains complex reasoning scenarios and broad semantic variation. These results show that HBLR offers strong efficiency across diverse tasks.

\subsection{Performance Across Reasoning Depths}
\begin{figure}[t]
    \centering
    \includegraphics[width=\linewidth]{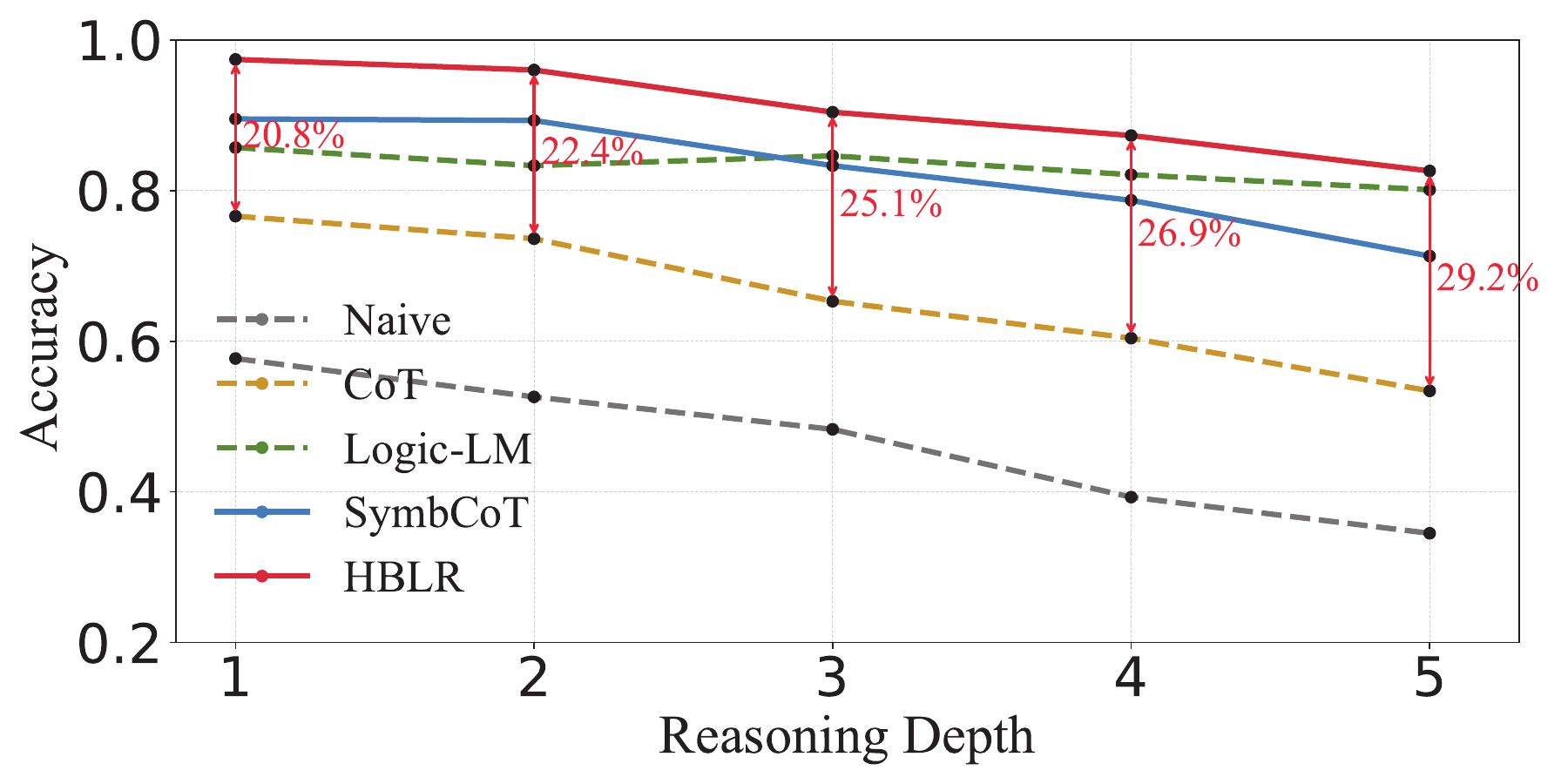}
    \caption{The effect of reasoning depth with GPT-4 on ProofWriter. The red double-headed arrow indicates our improvements over vanilla CoT.}
    \label{fig:zhexian}
\end{figure}
Having established the overall superiority of our method in direct comparisons, we further analyze its performance across different levels of reasoning depth. Intuitively, greater reasoning depth corresponds to higher problem complexity. As illustrated in Figure~\ref{fig:zhexian}, the performance gap between HBLR and CoT widens as reasoning depth increases, highlighting HBLR’s advantage in handling more challenging reasoning scenarios.
Notably, even at a depth of 5—the most complex setting in our evaluation—HBLR continues to achieve the highest performance among all methods.

\subsection{Impact of Stronger Backbone Models}

\begin{table}[t]
\centering
{
\fontsize{9}{10}\selectfont
\begin{tabular}{lccc}
\toprule
\textbf{Method} & \textbf{DeepSeek-V3} & \textbf{DeepSeek-R1} & $\Delta$ \\
\midrule
Logic-LM & 74.69 & 80.93 & +6.24 \\
SymbCoT  & 78.96 & 88.23 & +9.27 \\
HBLR     & 82.31 & 94.34 & +12.03 \\
\bottomrule
\end{tabular}
}
\caption{Performance improvements from DeepSeek-V3 to DeepSeek-R1 on different methods.}

\label{tab:ds-upgrade}
\end{table}

We compare the performance gains achieved when upgrading from DeepSeek-V3 to the more reasoning-capable DeepSeek-R1 model, as shown in Table~\ref{tab:ds-upgrade}. HBLR shows larger improvements than Logic-LM and SymbCoT, primarily because these translation-heavy methods benefit less from an upgrade that enhances reasoning ability rather than translation quality. In contrast, HBLR’s confidence-aware translation mitigates this bottleneck, allowing it to better exploit the improved reasoning capacity of DeepSeek-R1.
HBLR’s overall gain is smaller than that of Direct and CoT, as its baseline performance on DeepSeek-V3 is already strong, especially on ProntoQA, where accuracy is near saturation. Methods with lower initial performance naturally show larger absolute improvements under the upgrade. These results highlight HBLR’s adaptability and its strong potential in settings where model-side reasoning improvements are increasingly central.

\section{Conclusion}
This study presents HBLR, a hypothesis-driven backward reasoning framework for natural language logical reasoning. HBLR combines confidence-aware partial symbolic translation with human-inspired backward chaining to enhance precision and interpretability. It translates only high-confidence spans into formal logic based on structural and semantic cues, leaving uncertain content in natural language to balance symbolic rigor and flexibility. Reasoning begins from the hypothesis and works backward to validate supporting premises, aided by a reflection mechanism that enforces step-wise consistency. Across five benchmarks with diverse symbolic settings, HBLR consistently surpasses prior state-of-the-art methods in both accuracy and effectiveness.

\appendix

\section*{Acknowledgements}
This research was supported by grants from the National Natural Science Foundation of China (No. 62502486), the grants of Provincial Natural Science Foundation of Anhui Province (No. 2408085QF193), the Fundamental Research Funds for the Central Universities of China (No. WK2150110032).

\bibliography{aaai2026}

\end{document}